\documentclass[sigconf]{acmart}
\usepackage{multirow}
\usepackage{subcaption}
\usepackage{tabularx}

\AtBeginDocument{%
  \providecommand\BibTeX{{%
    \normalfont B\kern-0.5em{\scshape i\kern-0.25em b}\kern-0.8em\TeX}}}

\setcopyright{acmcopyright}
\copyrightyear{2018}
\acmYear{2018}
\acmDOI{XXXXXXX.XXXXXXX}

\acmConference[Conference acronym 'XX]{Make sure to enter the correct
  conference title from your rights confirmation email}{June 03--05,
  2018}{Woodstock, NY}
\acmPrice{15.00}
\acmISBN{978-1-4503-XXXX-X/18/06}

\begin{document}

\title{Explainable Hierarchical Urban Representation Learning for Commuting Flow Prediction}

\author{Mingfei Cai}
\affiliation{%
  \institution{Department of Civil Engineering, The University of Tokyo}
  \streetaddress{Komaba 4-6-1}
  \city{Meguro}
  \state{Tokyo}
  \country{Japan}
  \postcode{113-8656}
}
\email{mfcai@iis.u-tokyo.ac.jp}

\author{Yanbo Pang}
\affiliation{%
  \institution{Center for Spatial Information Science, The University of Tokyo}
  \streetaddress{Komaba 4-6-1}
  \city{Meguro}
  \state{Tokyo}
  \country{Japan}}
\email{pybdtc@csis.u-tokyo.ac.jp}

\author{Yoshihide Sekimoto}
\affiliation{%
 \institution{Center for Spatial Information Science, The University of Tokyo}
 \streetaddress{Komaba 4-6-1}
 \city{Meguro}
 \state{Tokyo}
 \country{Japan}}
\email{sekimoto@csis.u-tokyo.ac.jp}

\renewcommand{\shortauthors}{Cai et al.}

\begin{abstract}
  Commuting flow prediction is an essential task for municipal operations in the real world. Previous studies have revealed that it is feasible to estimate the commuting origin-destination (OD) demand within a city using multiple auxiliary data. However, most existing methods are not suitable to deal with a similar task at a large scale, namely within a prefecture or the whole nation, owing to the increased number of geographical units that need to be maintained. In addition, region representation learning is a universal approach for gaining urban knowledge for diverse metropolitan downstream tasks. Although many researchers have developed comprehensive frameworks to describe urban units from multi-source data, they have not clarified the relationship between the selected geographical elements. Furthermore, metropolitan areas naturally preserve ranked structures, like cities and their inclusive districts, which makes elucidating relations between cross-level urban units necessary. Therefore, we develop a heterogeneous graph-based model to generate meaningful region embeddings at multiple spatial resolutions for predicting different types of inter-level OD flows. To demonstrate the effectiveness of the proposed method, extensive experiments were conducted using real-world aggregated mobile phone datasets collected from Shizuoka Prefecture, Japan. The results indicate that our proposed model outperforms existing models in terms of a uniform urban structure. We extend the understanding of predicted results using reasonable explanations to enhance the credibility of the model.
\end{abstract}



\maketitle

\section{Introduction}\label{sec:intro}

It is crucial to reveal the metropolitan system that encompasses myriad elements, including but not limited to, road infrastructures and point of interests (POIs)~\citep{zheng2014urban}. Human mobility can serve as an interactive indicator for the state of the urban environment, and commuting flow is a suitable signal to understand general massive human mobility. Prediction of commuting flows generates a daily origin–destination (OD) matrix between different pairs of urban units from \textbf{socioeconomics features}. It is indispensable to note that forecasting future OD flows based on historical series~\citep{zou2021long} and predicting commuting flows from open-source regional attributes are two branches of OD-related studies~\citep{liu2020learning}. Considering the cost of input data and potential privacy issues~\citep{cai2022spatial}, the generation of commuting flows without expensive and sensitive movement histories is crucial. 

\begin{figure}
  \centering
  \begin{subfigure}[b]{0.99\linewidth}
    \centering
     \includegraphics[width=\linewidth]{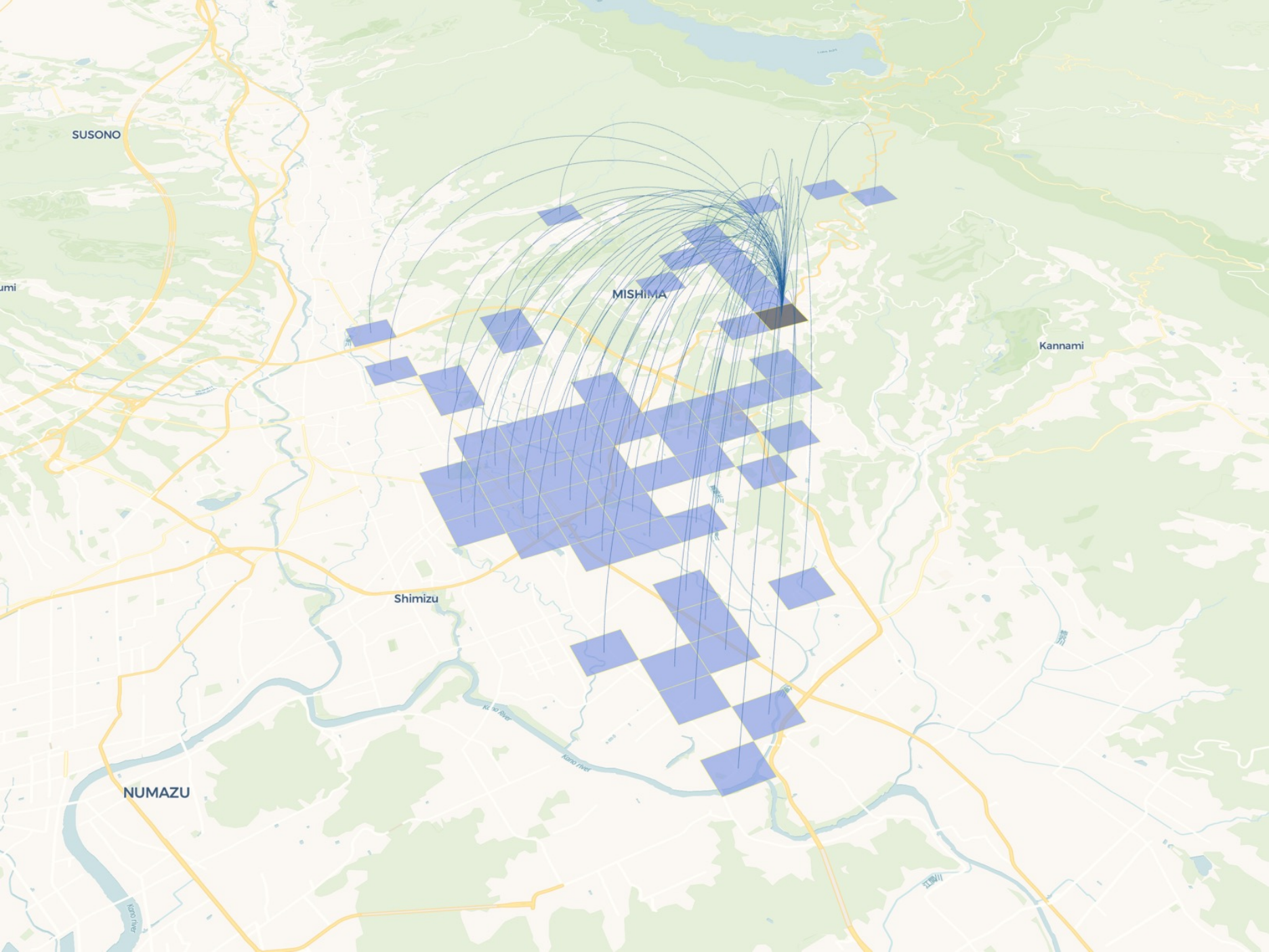}
     \caption{Intra-level commuting flows between grids and grids}
     \label{fig:m2m}
  \end{subfigure}
  \hfill
  \begin{subfigure}[b]{0.99\linewidth}
    \centering
     \includegraphics[width=\linewidth]{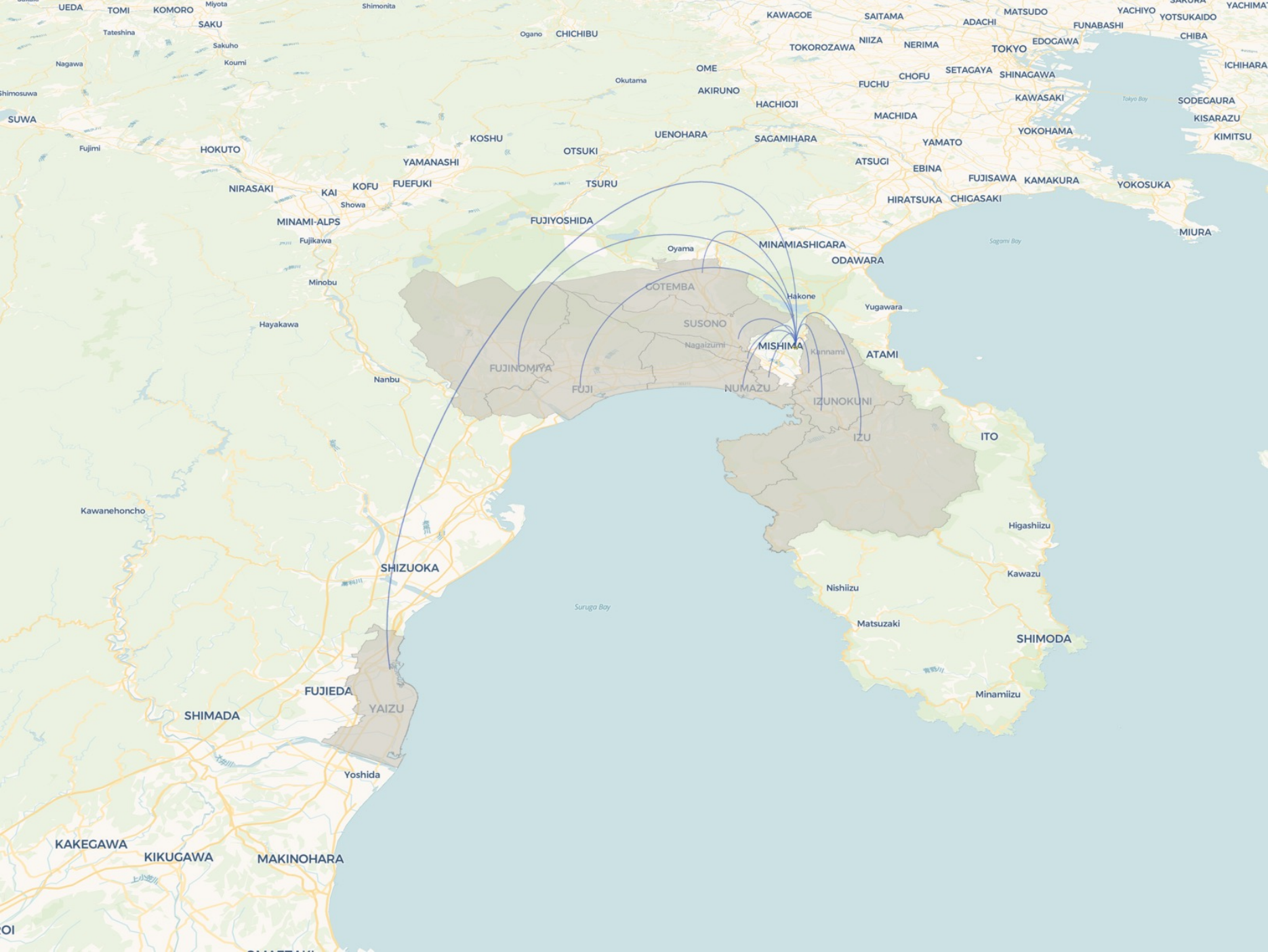}
     \caption{Inter-level commuting flows between cities and grids}
     \label{fig:c2m}
  \end{subfigure}
  \caption{Different types of hierarchical commuting flows. Subfigure~\ref{fig:m2m} indicates trip flows between grids and grids, while subfigure~\ref{fig:c2m} shows people flows between cities at higher levels and grids at lower levels.}
  \Description{Different types of hierarchical commuting flows}
  \label{fig:hierarchical_od}
\end{figure}

Predicting large-scale commuting flows is necessary and challenging, which suffers from out-of-distribution~\citep{zeng2022causal} and long-tailed-distribution problems. Most previous related studies conducted the task among uniform units, such as grids or census tracts, within a city, like New York~\citep{liu2020learning} or Beijing~\citep{shi2020predicting}. If the scope is broadened to the whole prefecture or nation, the model is not scalable to handle the exponential number of target units and quadratic number of relations~\citep{zhang2022dynamic}. It is even worse that a flat region structure cannot reveal meaningful patterns on a large scale. For instance, it is reasonable to estimate the people flow between two square grids in a local city, as indicated in figure~\ref{fig:m2m}, whereas it is of no use to describe such a flow across a long distance, such as two grids in two separate countries. The city planners are more interested in quantification of urban flows from a small grid to another cities or countries across the long distance. Therefore, it is critical to maintain different resolutions for depicting large-scale mobility patterns, as showed in figure~\ref{fig:c2m}. Fortunately, metropolitan areas naturally maintain ranked structures. Consistent with government policies, a country is divided into several prefectures, states, or provinces, and these subunits are further divided into cities. The cities include districts, census tracts, or grid areas, which are typically the smallest units in the national censuses of different countries. Hence, with such hierarchical divisions, employing the ranked property is viable to handle large-scale mobility predictions.

Additionally, to illuminate the mechanism of urban systems, it is significant to connect large-scale commuting flow predictions to the urban entities for a better comprehension. Urban representation learning is an effective approach for fully utilizing mobility information for users like governments. Some previous trials described areas numerically to benefit from criminal predictions~\citep{luo2022urban} and land usage classifications~\citep{zhou2023hierarchical}. However, most regional representation models exploited the same origins or destinations counts as mobility information to calculate likelihoods between locations~\citep{chan2023region}, which overlooked meaningful dynamic flow information. Furthermore, the representation of areas at a higher level like cities is more complicated than the simple addition of features from regions at a lower level. It is critical to gain deep insight into the spatial components of different levels to obtain a better summary of hierarchical urban areas.

Therefore, we propose a heterogeneous graph network-based model, \underline{Hi}erarchical \underline{Ur}ban \underline{Net}work (\textbf{HiUrNet}), to handle multiscale regional representation learning tasks based on commuting flow prediction. To address the aforementioned limitations and challenges, we collected multiple open-source data as the urban indicators and constructed a knowledge graph-like structure to indicate both static municipal and dynamic flow relationships. HiUrNet utilizes the attention mechanism to capture relationships parameterized by both urban unit types and semantics interactions. We further reconstruct the commuting flow matrices by learned multilevel urban unit embeddings, and conducted a multi-task learning task to deal intra-level and inter-level flows.

Moreover, explainable artificial intelligence aims for trustworthy outputs generated by machine learning algorithms. In this study, we follow the philosophy of the graph explaining algorithm~\citep{amara2022graphframex} to explain the prediction of HiUrNet from the viewpoint of the graph learning. This module offers a reasonable understanding for important grids that mainly determine the description of the upper-level cities. Such knowledge can provide an insightful view for policy-makers to maintain facilities to increase the attractions of the urban areas. The main contributions of this study are as follows:

$\bullet$ We implemented HiUrNet, a heterogeneous graph-based model, to learn a hierarchical urban unit embedding using multi-source open information. HiUrNet handles different types of urban relation triplets by type-specific attention scores, and fuses static and dynamic neighboring messages throughout a relation-aware aggregation.

$\bullet$ We utilized the produced embeddings for predicting large-scale hierarchical commuting flows in a multi-task learning scene. Moreover, we used an explanatory module to explain the manner in which the city embeddings are summarized throughout neighboring message pass process.

$\bullet$ We conducted extensive experiments in a Japanese prefecture, and the results indicated that our proposed model offered satisfactory and explainable commuting OD flow predictions. 

To the best of our knowledge, this is the first study to generate multilevel urban unit representations for prediction of hierarchical commuting flows.

\section{Related Work}

\subsection{Commuting Flow Prediction}

Commuting flow prediction~\citep{zhou2023reconstruction}~\citep{wang2023generating}~\citep{rong2021inferring} is a demanding task in transportation and urban planning. The task provide insights for massive people flows under geo-contextual scenarios. Note that commuting flow prediction is different from the OD forecasting problem~\citep{huang2023odformer}~\citep{wang2019origin}, which outputs the same OD matrices but uses OD time series as input~\citep{liang2021fine}. This different problem definition cannot consider areas lacking sufficient history data. Such tasks focused on time variation factors rather than the region itself, lacking the ability to be reused in other urban downstream tasks. Some researchers have attempted to utilize socioeconomics data to predict commuting flows~\citep{simini2021deep}~\citep{liu2020learning}. Regardless of predictions of human mobility between census tracts or mesh grids, they adopted default settings that origins and destinations were at the same level. Unfortunately, these models could not make predictions at a large scale, such as within the entire country, owing to the explosive number of target units. The proposed predictions could not also illustrate complex human travel behaviors. Moreover, the data distribution of large-scale human mobility, including multiple transportation modes, tends to be more complicated and skewed and requires a more powerful model to provide accurate predictions. In this study, we utilize hierarchical city structures to reduce the unnecessary number of urban units and derive more useful moving patterns for the government to analyze the related policies.

\subsection{Multilevel Regional Representation Learning}

Representation learning is an effective method for extracting meaningful information from multiple data sources without requiring arduous feature engineering. It is suitable for urban scenes owing to the existence of multi-modal data and entities. In general, regional representation learning is to use numerical expressions to describe areas. Some past trials that utilized image-like embedding learning~\citep{jiang2021transfer} only considered Euclidean distance for different entities. Recent studies~\citep{luo2022urban} have often implemented contrastive learning using static features to explore embedding spaces in an unsupervised manner~\citep{huang2023learning}. Most of these studies did not consider dynamic mobility-related patterns or simply used the existence of flow rather than volumes as part of the construction of the knowledge graph~\citep{zhang2021multi}, which renders the model ineffective in handling spatial information. \citet{luo2022urban} constructed a graph based on the neighborhood, cosine similarity of POIs, inflows/outflows and footprints~\citep{li2023urban}, and constructed graphs in multiple views and fused them to generate the final representations~\citep{wu2022multi}. The downstream tasks are often set as the crime prediction and land-usage classification for evaluating embedding quality. However, although the learned embeddings achieve good performance on the node-like tasks, the performance tend to deteriorate in edge-like tasks. Regarding hierarchical urban structures, attempts to elaborate spatial~\citep{mai2020multi} or temporal~\citep{zhang2022dynamic} representations at different levels have been made. For example, \citet{wu2020learning} proposed a framework for clustering road networks and functional areas. These models can generate hierarchical region embeddings. However, the hierarchy of cities is typically pre-determined by governments using different policies. Developing a framework for obtaining knowledge about structural administrative regions is critical. In this study, we utilized static urban attributes and dynamic mobility information to generate more robust regional representations.

\subsection{Explainable Urban Computing}

The black-box nature of deep learning-based models can hamper the usage of the proposed results without fidelity, specifically in real-world applications such as urban data mining. Therefore, explainable deep learning models have recently received considerable attention. \citet{simini2021deep} implemented Shapely additive explanations to analyze contribution of input geographic features based on game theory. \citet{zhou2024explainable} utilized Monte Carlo sampling to calculate Shapely values of inputted features. However, these studies do not explain the predicted phenomenon based on both urban feature semantics meanings and syntactic relationships between different urban entities. 

\section{Preliminaries}

Here, we define the key concepts and formulate the problem. The study aims to learn the hierarchical representations of urban units for commuting flow predictions. Table~\ref{tab:symbol} summarizes main symbols used in this study. Note that we use uppercase letters to denote the set, and the lowercase ones to denote the entities in the set, respectively.

\begin{table}[htbp]
 \label{tab:symbol}
 \centering
 \caption{The summary of main symbols in this study.}
 \begin{tabular}{cl}\toprule
    Symbol&Description\\
    \cmidrule(lr){1-2} 
    $\mathbf{U_c}$, $\mathbf{U_m}$, $u^m$, $u^c$&urban units as cities or mesh grids\\ 
    $N_c$, $N_m$ & The number of urban units\\
    $\mathbf{X_m}$ & The urban indicator matrix of mesh grids\\
    $\mathcal{G},\mathcal{V},\mathcal{E}$ &The urban graph, vertices and edges\\
    $D$& The embedding size\\
    $\mathcal{F}$& The mapping function for urban embeddings\\
    $\mathcal{H}$& The urban unit embedding\\
    $\mathbf{A}^\phi$&The matrix for commuting flow in type $\phi$\\
 \bottomrule
 \end{tabular}

 \label{tab:symbols}
\end{table}

{\bfseries Definition 1 (urban units)}. Urban units $\mathbf{U}$ are geographical elements with the attributes of urban environments. In this study, there are two types of urban units: mesh grids and cities. Mesh grids $\mathbf{U_m}=\{u_1^m,u_2^m,…,u_{N_m}^m\}$ are non-overlapping areas of equal size (e.g., $500m \times 500m$), attached with unique socioeconomics indicators $\mathbf{X_m} \in \mathbb{R}^{N_m \times D}$, describing the $D$ dimensional attributes of the $N_m$ mesh grids. We adopt the division method from the Statistics Bureau of Japan~\citep{web:mesh}. Cities $\mathbf{U_c}=\{u_1^c,u_2^c,…,u_{N_c}^c\}$are the units that include a set of mesh grids. The shapes of cities are diverse as defined by the government. The inclusion relationship between cities and mesh grids is indicated by a mapping function $\omega(\textrm{inclusion}):\mathbf{U_c} \rightarrow \textrm{List}(\mathbf{U_m})$.

We consider reconstructing commuting flows between pairs of urban units using OD matrices. An OD matrix is a type of mobility data used to describe the flow of many people. OD matrices aggregate individual flows and avoid privacy tolerance to efficiently indicate macroscopic human mobility, thus chosen as the predicted target.

{\bfseries Definition 2: (hierarchical commuting OD matrix)}. In a hierarchical OD background, each OD matrix is associated with a type defined by the origin and destination. For the commuting OD matrix with a specific type $\mathbf{A}^\phi~\in~\mathbb{R}^{N_{\mathbf{U_i}} \times N_{\mathbf{U_j}}},\mathbf{U_i},\mathbf{U_j} \in \{\mathbf{U_m},\mathbf{U_c}\}$, the row denotes the origin with the type $R_i$, and the column denotes the destination with the type $R_j$. The value of each cell $a_{i,j}$ indicates the number of people traveling from origin $r_i$ to destination $r_j$. 

We can construct hierarchical urban graphs from the OD matrices to simultaneously consider the semantics of the urban units and the topological structure information.

{\bfseries Definition 3: (hierarchical urban graph)}. The hierarchical urban graph is formulated as a directed graph $\mathcal{G}=(\mathcal{V},\mathcal{E})$, where $\mathcal{V}$ represents the vertices (urban units) and $\mathcal{E}$ is the edge connecting the two vertices (inclusion and flow relations). The vertex and the edge have the relative types. In general, the edges include dynamic flow types and static inclusion types, which are illustrated by the OD matrices and city grid inclusion mapping functions. 

{\bfseries Problem Definition}: Given a hierarchical urban graph $\mathcal{G}$ with grid indicators $\mathbf{X_m}$, the model learns a function $\mathcal{F} \textrm{:} \mathcal{V}\rightarrow\mathcal{H}\in\mathbb{R}^d$ that maps urban units to $d$ dimensional embeddings for reconstructing the commuting flow matrices $\mathbf{A}^\phi$ in different types.

\section{Methodology}

\begin{figure*}[htb]
  \centering
  \includegraphics[width=0.99\textwidth]{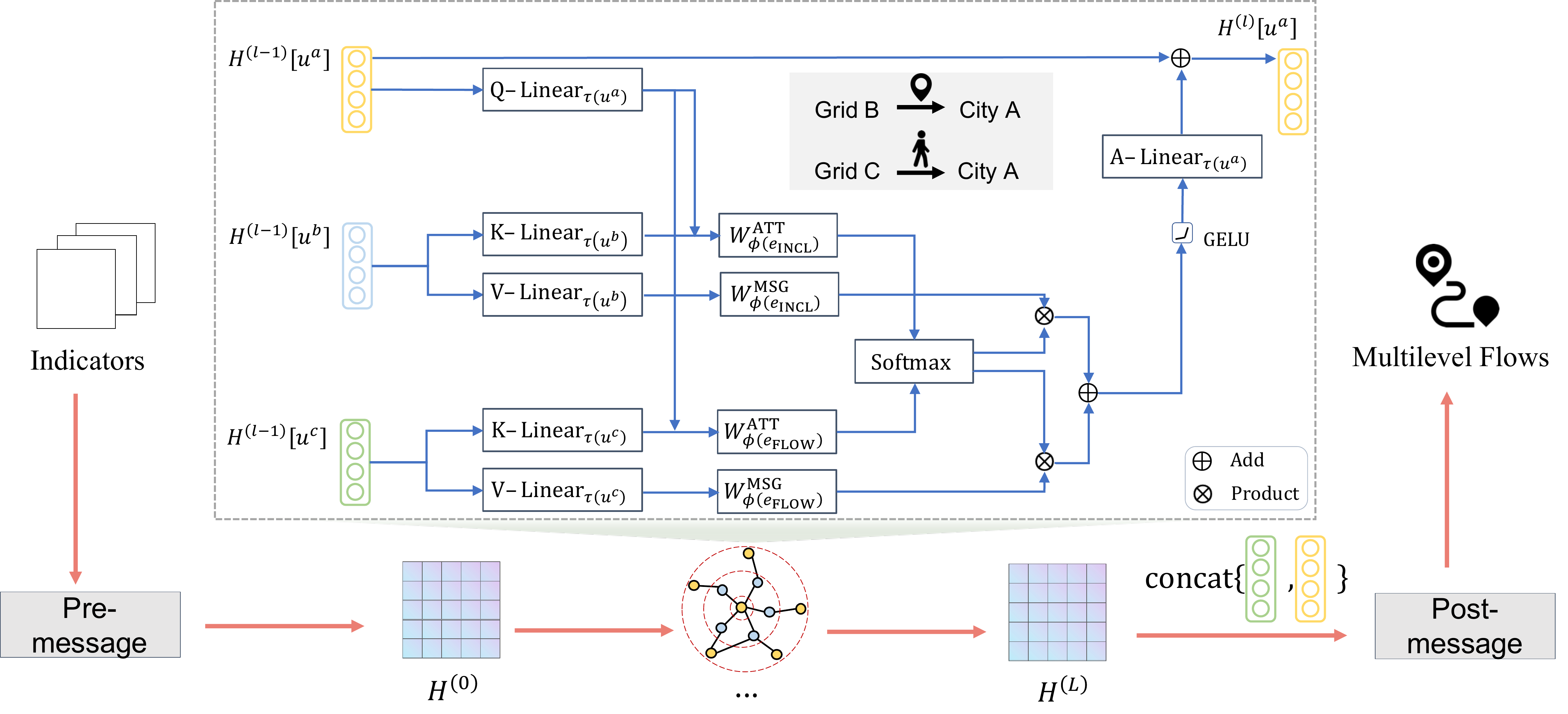}
  \caption{The overview of Hierarchical Urban Network model. In general, the framework includes the pre-message process for initialization of embeddings using urban indicators, urban relation-aware graph convolution and the post-message process for multilevel flow reconstruction. We take a typical urban subgraph for the instance to introduce our proposed message passing method. There are three urban units and two relations, that is, grid B locates at city A and grid C has flows to city A.}
  \label{fig:outline}
\end{figure*}

Here, we provide details of the proposed framework, including the hierarchical urban graph construction, model architecture design, and explanation generation.

\subsection{Hierarchical Urban Graph Construction}

This study aims to reveal the complicated relations between different spatial units and urban indicators; thus, we modeled the relationship of diverse components using graph structures. We represented a regional graph as a heterogeneous graph with different types of nodes and edges. The nodes included city and grid types, where the number of cities was significantly fewer than the number of grids. Because the people flow signals are the main targets used for extracting meaning region embeddings, we constructed three flow-type edges considering different combinations of origins and destinations. Additionally, to depict the relationship between cities and mesh grids based on administrative division from the government, we designed include-type and the inverse in-type edges. The details of the hierarchical urban graphs are presented in the table~\ref{tab:graph}.

\subsection{Hierarchical Urban Network}

In this subsection, we introduce the details of the proposed HiUrNet. Figure~\ref{fig:outline} provides an overview of this framework. We used graph neural network (GNN) with a multi-head heterogeneous attention mechanism to perform hierarchical urban unit representation learning. Previous studies~\citep{cai2022spatial} have often used backbones such as graph attention network (GAT) models, in which a uniform attention coefficient matrix is calculated to weight different neighbors. However, this design is not appropriate for handling complex patterns and mutual effects between different edge types. In relation to the diverse regional types and their inter-regional relationships, distinct attention coefficients should be computed to discern associated trends. Inspired by the heterogeneous graph transformer (HGT)~\citep{hu2020heterogeneous}, we used relations between urban units to parameterize the attention matrices and applied type-specific aggregation on information fusion from the neighborhood. 

First, the model leverages multisource open data for initializing the embeddings of grids, while cities are affiliated with one-hot vectors owing to their larger sizes and smaller numbers. In general, updated embeddings originate from two parts: direct information from the source node and the attention factor. Under GNN scenario, the center node accepts messages from neighboring nodes to update its representation. The message from the source node is calculated by

\begin{equation}
    \mathbf{MSG}\left(u^s,e,u^t\right)=\underset{i \in [1,h]}{\parallel}\textrm{M-Linear}_{\tau(u^s)}^{i}(H^{(l-1)}[u^s])W_{\phi(e)}^{\textrm{MSG}}
\end{equation}

where $u^s$ and $u^t$ are the source and target urban unit vertices, respectively, and $e$ is the edge connecting these two nodes. We concatenate multi-heads of message to formulate the final result. $\tau(\cdot)$ determines the vertex type and $\phi(\cdot)$ specifies the edge type. $\textrm{M-Linear}_{\tau(u^s)}^{i}$ is the linear projection for message parameterized by the type of source urban units, while $W_{\phi(e)}^{\textrm{MSG}}$ is the edge-based matrix to incorporate static inclusion and dynamic flow information. $H^{(l-1)}$ is the output of the $(l-1)\textrm{-th}$ layer, which is the input for the $l\textrm{-th}$ layer. 

For attention scores, different key and value matrices are generated from the source nodes for different edge types. The target node serves as a query and generates a query matrix. Thus, the attention matrix of the $i^{th}$ head is formulated by

\begin{equation}
    \textrm{ATT}^i(u^s,e,u^t)=(K^i(u^s)W_{\phi(e)}^{\textrm{ATT}}Q^i(u^t)^T) \cdot \frac{\mu_{\langle \tau(u^s),\phi(\mathbf{e}), \tau(u^t) \rangle}}{\sqrt{d}}
\end{equation}

$W_{\phi(e)}^{\textrm{ATT}}$ is similar to $W_{\phi(e)}^{\textrm{MSG}}$ but for the attention calculation. $\mu_{\langle \cdot \rangle}$ is the adaptive scaling to the attention, and $d$ is the vector dimension. Specifically, the key from the source node and query from the target node are calculated as follows:

\begin{equation}
    K^i(u^s)=\textrm{K-Linear}_{\tau(u^s)}^{i}(H^{(l-1)}[u^s])
\end{equation}
\begin{equation}
    Q^i(u^t)=\textrm{Q-Linear}_{\tau(u^t)}^{i}(H^{(l-1)}[u^t])
\end{equation}

$\textrm{K-Linear}$ and $\textrm{Q-Linear}$ are the linear projection layers for the key and value, respectively. The attention factor is then calculated as follows:

\begin{equation}
    \mathbf{ATT}=\underset{\forall u^s \in N(u^t)}{\mathrm{Softmax}}(\underset{i \in [1,h]}{\parallel} \textrm{ATT}^i(u^s,e,u^t))
\end{equation}

Note that edges in one type share the same linear transformation layers in equations. The message then aggregates the information with different edge types. In general, the edge types encompass urban-flow and municipal-include types, thereby distinctly indicating inclusion and mobility semantics. These settings can also alleviate the long-range node dependency of the two virtual nodes. The temporary updated embeddings can be represented as

\begin{equation}
    \widetilde{H}^{(l)}[u^t]=\underset{\forall u^s \in N(u^t)}{\oplus} (\mathbf{ATT}_{(u^s,e,u^t)} \cdot \mathbf{MSG}_{(u^s,e,u^t)})
\end{equation}

Finally, the target urban unit embedding is updated by adding the residual part

\begin{equation}
    H^{(l)}[u^t]=\textrm{A-Linear}_{\tau(u^t)}(\sigma(\widetilde{H}^{(l)}[u^t])+H^{(l-1)}[u^t])
\end{equation}

$\textrm{A-Linear}$ is the linear projection layers for the aggregation. Thereafter, we applied the generated representations for reconstructing the large-scale commuting flows. We derived three types of commuting flow matrix predictions: city-to-grid, grid-to-city, and grid-to-grid as multiple tasks that share the same bottom layers. The loss is defined as

\begin{equation}
    \mathcal{L}_{\textrm{mtl}} = \sum {\omega_i} \cdot L_i
\end{equation}

The city-to-grid and grid-to-city trips were similar and set with the same task weights. After extracting the region embeddings at the city and mesh grid levels, the urban unit pairs were passed to different decoders. 

For the loss for a specific task $t_k$, we adopt focal L2 regression loss to tackle unbalanced data regression problems~\cite{yang2021delving}. 

\begin{equation}
    \mathcal{L}_{t_k} = \frac{1}{n}\sum_{i}{(2\sigma(|\beta e_i|)-1)^\gamma e_i^2} 
\end{equation}

where $n$ denotes the number of edges. $\beta$ and $\gamma$ are the hyperparameters selected by the grid search, where $\beta = 0.2$ and $\gamma = 1$. $e_i$ is the difference between the prediction and the ground-truth labels of edge $i$ and $\sigma(\cdot)$ is the sigmoid function.

\subsection{Regional Summary Analysis}

Although learning embeddings for accurate predictions of commuting flows are necessary, the explanation of reasonable predictions is essential. Because a city consists of multiple mesh grids, we must determine the grids that are more significant for describing the entire city. Inspired by the Captum module~\citep{kokhlikyan2020captum}, which is widely used to explain predictions from the deep learning models, we conducted a city summary analysis based on the generated embeddings. We adopted an integrated gradient as the axiomatic attribution method~\citep{sundararajan2017axiomatic}.

Specifically, we selected a target city after the training of the model. Thereafter urban unit pairs with the largest inter-level flows with the target city were extracted as the target edges to explain. The integrated gradients method took the target edges as inputs and calculated the attribution by running the model. The masks for nodes and edges were generated based on the attribution for the hierarchical city graph. Throughout these masks, the regional summary analysis was conducted considering urban indicators and city-grid inclusion relationships to elucidate the essential indicators for accurate commuting flow predictions. This analysis can also illustrate significant mesh grids to describe the related upper-level city.

\section{Experiments}

Here, we present the details of the experiment, including the data description, experimental settings, and selection of baseline models. The experiment includes two parts: the commuting flow predictions and the explanation for the proposed predictions. The code is available at \url{https://github.com/shishixuezi/HiUrNet}.

\subsection{Experimental Setup}

In this subsection, we describe the experimental settings used to validate the model. The experiment was performed in Shizuoka Prefecture, Central Japan. Shizuoka Prefecture covers $7777.43 km^2$ and has a population over three million people. In previous studies, researchers often chose one city, such as New York City, which contains hundreds of subareas, such as grids or census tracts. In this study, we selected an area in the higher hierarchy, namely the prefecture. These areas cover a larger scope and maintain more complicated and meaningful mobility patterns that require further exploration.

\subsection{Dataset}

Here, we present the details of the datasets. The datasets were related to commuting flows, grid inclusion, and grid indicators.

\subsubsection{Commuting Flows}

This study used the National Move Statistics as the people movement source data, provided by the SoftBank Group Corporation, one of the biggest telecom companies in Japan. The data were preprocessed from the raw call details records data. The flows were determined to be commuting trip pairs according to the distribution of the signal receiver towers and up-sampling rates. A summary of the data for the target area is presented in table~\ref{tab:graph}.

\begin{table}
  \caption{Summary of the hierarchical urban graph}
  \label{tab:graph}
  \begin{tabular}{cc}
    \toprule
    \bf{Node/Edge} Type & \bf{Num}\\
    \midrule
    mesh & 37108\\
    city & 43\\
    (mesh, m2c, city) & 62684\\
    (city, c2m, mesh) & 62893\\
    (mesh, m2m, mesh) & 320905\\
    (city, includes, mesh)& 37108\\
  \bottomrule
\end{tabular}
\end{table}

\subsubsection{Grid Indicators}

This study aims to multiple open-source datasets for learning region descriptions. We used several grid indicators in different categories to initialize the embeddings of the mesh grid. Specifically, residential population~\footnote{\url{https://nlftp.mlit.go.jp/ksj/gml/datalist/KsjTmplt-mesh500h30.html\#prefecture22}}, road density~\footnote{\url{https://nlftp.mlit.go.jp/ksj/gml/datalist/KsjTmplt-N04.html}}, railway users~\footnote{\url{https://nlftp.mlit.go.jp/ksj/gml/datalist/KsjTmplt-S12-v2_7.html}}, and POIs distribution~\footnote{\url{https://www.e-stat.go.jp/gis/statmap-search?page=1&type=1&toukeiCode=00200553\&toukeiYear=2016&aggregateUnit=H\&serveyId=H002005112016\&statsId=T000918}} were selected as grid indicators for human mobility scenes. Residents are the main source of people flow; thus, they should not be neglected as the urban indicators. Moreover, it can infer the functionality of regions at a certain degree. For instance, if a grid has a small residential population but a large people flow, it could be related to a business district. In addition, road density and railway users can elucidate the convenience of traffic infrastructure. As general commuting flows are the prediction targets, such indicators are advantageous in this study over others that consider only a single traffic mode. Note that the distribution of facilities in an area is an essential description of the trip’s purpose. Therefore, we included POIs distribution as another indicator data source. These data summarize the number of facilities in diverse categories. The data source for the grid indicators was the census of the Ministry of Land, Infrastructure, Transport and Tourism of Japan (MLIT) and Statistics Bureau of Japan. Table~\ref{tab:indicators} presents the details of the data sources.

\begin{table}
  \caption{Summary of the grid indicators}
  \label{tab:indicators}
  \begin{tabularx}{0.99\columnwidth}{>{\centering}c
>{\centering}c
>{\centering\arraybackslash}X}
    \toprule
    \bf{Feature Categories}&\bf{\#Features}&\bf{Contents}\\
    \midrule
    Road Densities &24&Road number and density of different widths for mesh \\
    Facilities (POIs)&17&Number of facilities and employees in different industrial categories \\
    Grid Population &1 & Night population distribution \\
    Railway Users &1 & Number of annual railway station users \\
    \midrule
    Total&43&\\
    \bottomrule
\end{tabularx}
\end{table}

\subsubsection{Baseline Models}

We compared the proposed model with several baseline models. We chose models ranging from classical models to state-of-the-art deep learning models. We selected typical urban representation learning models and commuting flow prediction models considering the motivation of this study in section~\ref{sec:intro}.

\begin{itemize}
    \item The gravity model is a classical approach inspired by the gravity law in physics. It uses the static population distribution and spatial distance to estimate the volume of flows. We adopted the generalized form of the gravity model and used simple multi-layer perceptron (MLP) layers to optimize the parameters of the different functions.
    \begin{equation}
        T_{ij}=f_i(P_i)f_j(P_j)f_d(d_ij)
    \end{equation}
    \item The decision tree is a nonparametric tree-based model that can be used in regression problems. The data features were utilized for the inference of simple decision rules to provide an accurate prediction.
    \item The random forest uses several decision tree regressors on subsets of the target data. The average of results is utilized to improve the accuracy and avoid overfitting patterns.
    \item The gradient boosting model is another type of tree-based model commonly used in regression analyses. It assembles several weak models to enhance the performance.
    \item The MLP model is often used as a basic brick for complicated deep learning architecture. It can be seen as a decoder-only variants of our proposed model.
    \item The MVURE model~\citep{zhang2021multi} is a typical static urban representation learning model. Note that the model utilized probability distribution of inflow/outflow as mobility input, which is the extra prior knowledge compared with our proposed model. Because the downstream tasks of MVURE are regressions for check-in and crime predictions for single urban units, we attach extra MLP layers for implementation of pair-wise regression to improve the performance.
    \item The deep gravity model~\citep{simini2021deep} assumes that the total outflows for an origin is determined and use a deep neural network to estimate the statistical distribution for each destination. The partial data leakage problem owing to the prior mobility statistics is similar to MVURE.
    \item The geo-contextual multitask embedding learner (GMEL) model is a state-of-the-art model used to estimate homogeneous commuting flows using multiple urban indicators~\citep{liu2020learning}. Because this model considers the flat structure of a city, we directly treat the city and mesh grid as the same.  
    \item $\textrm{HiUrNet}$ (proposed) is the hierarchical urban network model backboned by integrate attention mechanisms, similar to those adopted in HGT~\citep{hu2020heterogeneous}. The backboned GNN uses edge types to parameterize the attention mechanisms. 
\end{itemize}

As the baseline models are naturally unable to consider the hierarchical structure, we manually concatenated the indicator vectors of the origin and the destination as the input. For the city, we used the sum of the indicators from all grids it contains as the input for every baseline model. Note that although there are other state-of-the-art representation learning models, we only choose the MVURE owing to the serious mobility data leakage and difficult model implementation. For instance, MGFN~\citep{wu2022multi} adopted time-series mobility data for graph fusion. Such inputs are more costly than our proposed commuting flows. HREP~\citep{zhou2023heterogeneous} implemented prompts based on different downstream tasks to improve the performance, while similar prompts for commuting flows are difficult to define.

\subsubsection{Evaluation Metrics}\label{sec:metrics}

We measured the performance of the models with the following metrics for regression problems:
\begin{itemize}
    \item Root mean squared error (\textrm{RMSE}). The lower RMSE value is preferred.
    \begin{equation}
    \textrm{RMSE}(\hat{y_i},y_i) = \sqrt{\frac{\Sigma_{i=1}^{N}(\hat{y_i} -y_i)^2}{N}}
    \end{equation}
    \item Mean average error (\textrm{MAE}). The lower MAE value is preferred.
    \begin{equation}
    \textrm{MAE}(\hat{y_i},y_i) = \frac{\Sigma_{i=1}^{N}|\hat{y_i} -y_i|}{N}
    \end{equation}
    \item Pearson Correlation of Coefficient (\textrm{PCC}). The higher PCC value is preferred.
    \begin{equation}
    \textrm{PCC}(\hat{y_i},y_i) = \frac{\Sigma xy - \Sigma x \Sigma y}
    {\sqrt{[\Sigma x^2 - (\Sigma x)^2][\Sigma y^2 - (\Sigma y)^2]}}
\end{equation}
\end{itemize}

\subsection{Results and Discussion}

\begin{table*}[htbp]
\small
  \caption{Results of experiments. The results with the best performance are indicated by underlines.}
  \setlength{\tabcolsep}{8.2pt}
  \begin{tabular}{l ccc ccc ccc}
    \toprule 
    \multirow{2}{*}{\bf{Model}} & \multicolumn{3}{c}{\bf{Grid to Grid}} & \multicolumn{3}{c}{\bf{Grid to City}} & \multicolumn{3}{c}{\bf{City to Grid}} \\
    \cmidrule(lr){2-4}\cmidrule(lr){5-7}\cmidrule(lr){8-10}
                                & RMSE & MAE & PCC & RMSE & MAE & PCC & RMSE & MAE & PCC \\    
    \midrule
    \textsc{Gravity model} & $136.61$&$54.09$&$0.00$&
         $385.42$&$128.10$&$0.00$&
         $390.06$&$127.56$&$0.00$ \\
    \textsc{Decision tree} & $199.23$&$54.09$&$0.46$&
         $319.98$&$90.34$&$0.51$&
         $337.71$&$93.11$&$0.51$ \\
    \textsc{Random forest} & $159.63$&$44.67$&$0.24$&
         $263.20$&$81.59$&$0.45$&
         $273.85$&$81.21$&$0.40$ \\
    \textsc{Gradient boosting} & $158.26$&$43.89$&$0.39$&
         $255.27$&$77.61$&$0.56$&
         $265.55$&$76.84$&$0.54$ \\
    \textsc{MLP} & $158.81$&$47.08$&$0.36$&
          $261.64$&$91.99$&$0.50$&
          $273.22$&$98.60$&$0.50$ \\   
    \textsc{MVURE}~\citep{zhang2021multi} & $151.41$&$41.12$&$0.74$&
          $126.88$&$38.36$&$0.68$&
          $120.97$&$39.56$&$0.58$ \\
    \textsc{Deep gravity}~\citep{simini2021deep}& $134.29$&$44.12$&$0.07$&
          $258.89$&$90.34$&$0.54$&
          $196.22$&$60.83$&$0.54$ \\
    \textsc{GMEL}~\citep{liu2020learning} & $122.96$&$44.15$&$0.10$&
          $328.84$&$84.13$&$0.22$&
          $318.23$&$80.54$&$0.22$ \\
    \midrule
    $\textsc{HiUrNet}$ & $\underline{93.76}$&$\underline{29.94}$&$\underline{0.76}$&
          $\underline{89.24}$&$\underline{28.74}$&$\underline{0.88}$&
          $\underline{106.41}$&$\underline{31.70}$&$\underline{0.91}$ \\
    \bottomrule
  \end{tabular}
  \label{table:results}
\end{table*}

\begin{figure*} [htb]
  \centering
  \includegraphics[width=0.99\textwidth]{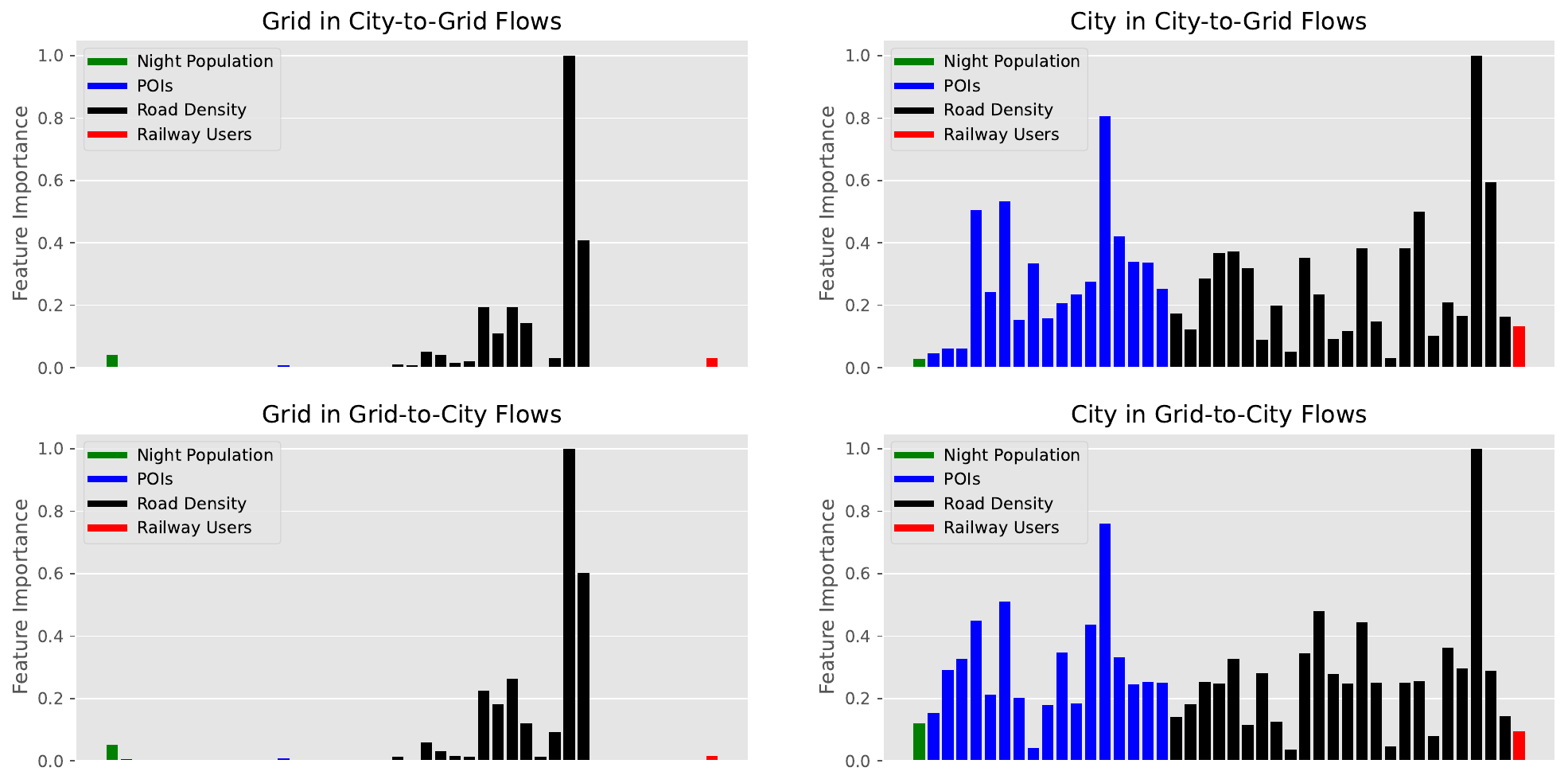}
  \caption{Feature importance analysis. For both city and grid embeddings, road network had a more significant impact than the other categories. The distribution of POIs was essential to calculate city representations, while the influence of facilities could be neglected for grid embeddings.}
  \label{fig:feature}
\end{figure*}

The model was constructed using the PyTorch geometric library~\citep{Fey/Lenssen/2019}. The ratio of training, validation, and test sets was set as $8:1:1$. For important hyperparameters, we decided the embedding dimensions as 128 for HiUrNet. The number of hidden channels was 256. We also implemented multi-head attention over graph convolution for stabilization. Specifically, we set the number of heads to eight for HiUrNet. Furthermore, we used the Adam optimizer for training. We conducted experiments on an Amazon p3.2xlarge instance with one NVIDIA V100 GPU, 8vCPUs, and 61 GiB of host memory.

\subsubsection{Commuting Flows Reconstruction}

We conducted experiments for all the baselines and the proposed model. Table~\ref{table:results} exhibits the summary of the performances. Regarding all the metrics mentioned in section~\ref{sec:metrics}, the proposed model achieved the best performance when considering different types of commuting flows.

Specifically, classical models, such as gravity models, cannot optimize predictions of massive pair data at a large scale. In addition, the universe regression models were ineffective in capturing spatial relationships for imbalanced commuting flows. The performance was worse for predictions of commuting flows between cities and mesh grids owing to the neglect of hierarchical urban structures. Urban representation models can perform well to portray urban units. However, the performance of these models cannot perform as well as they did in single-entity-related downstream tasks like crime predictions in pairwise predictions, as these models generally learned embeddings from likelihood through contrastive losses. Although the GMEL model performed well in reconstructing commuting flow matrices for taxis in New York City, it cannot provide a satisfying consequence for general commuting flows with several transportation modes. These results indicate the necessity of considering mutual effects between different spatial relationships.

\subsubsection{Indicator Importance Analysis}

Understanding the indicators is indispensable when utilizing multi-source urban data. Therefore, we conducted a sensitivity analysis of the different grid indicators. Figure~\ref{fig:feature} illustrates the importance normalized between zero and one regarding two types of inter-level people flows. For both city and grid embeddings, road network had a significant impact than the other categories. As road links are the direct medium for traffic flows, these densities can directly determine the variance in people flow. Moreover, there was a significant gap between cities and grids regarding POIs. The distribution of POIs was essential to calculate city representations, which was similar to the road densities. However, the influence of facilities could be neglected for grid embeddings. The fine-grained mesh grids covered minor areas; thus, the distribution of facilities was not a universe signal for people flows. 
In addition, the grid tenants were not negligible according to the analysis. These indicators also affect the number of people traveling from or to the target grids. The railway users were more significant for cities owing to the sparse distribution of railways.

\subsubsection{Summary Analysis}

\begin{figure}
  \centering
  \includegraphics[width=\columnwidth]{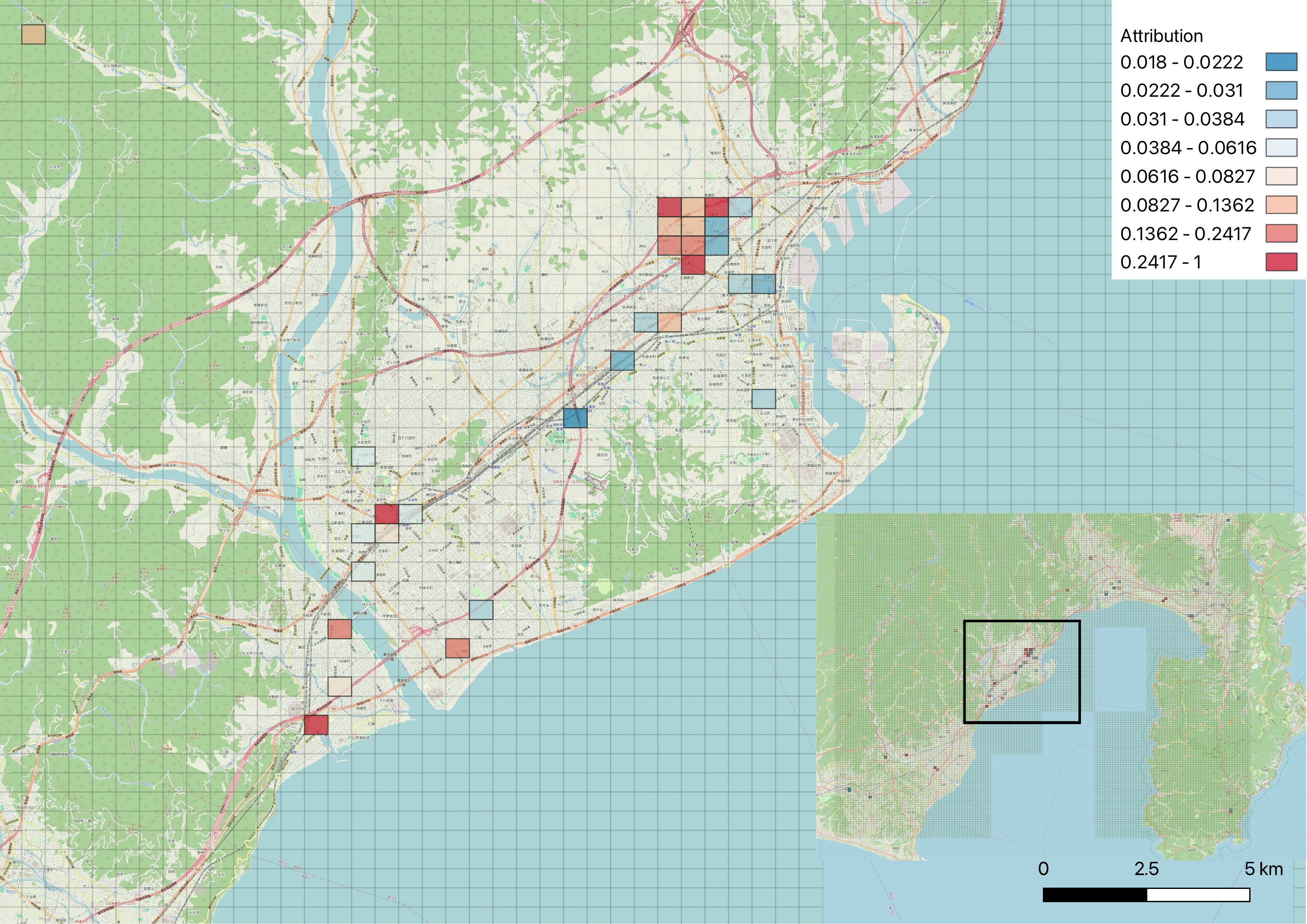}
  \caption{Grid attributions for city embeddings of Aoi Ward, Shizuoka City. The color of bars denotes the type of grid indicators and the length of the bar indicates the importance.}
  \label{fig:attribution}
\end{figure}

In the setting of the proposed HiUrNet model, the embeddings of cities are not simply initialized from the sum of its belonging grids. As the city embeddings are initialized randomly and are purely summarized from the included grid embeddings, we designed the summary analysis to show how different grids affect the city embeddings. Figure~\ref{fig:attribution} takes the Aoi Ward from Shizuoka City as an instance. The illustration elucidates that the city embedding of the Aoi Ward are mainly determined by populated grids in the central city along with those in the rural parts. The grids near the main branch of the road system tend to have a more important role for the comprehensive summary for the upper-level city representations, which meets the common knowledge that people flow has a close relationship with the urban facilities.

\subsection{Ablation Study}

Here, we conduct the ablation study to test the effect of our proposed model settings.

\subsubsection{Edge Types for Message Passing}

\begin{table*}
\small
\centering\centering
\setlength{\tabcolsep}{12.2pt}
\caption{Performance for different types of message passing edges in the proposed model. The best result is denoted by the underline.} 
\begin{tabular}{p{2cm}ccccccccc}
\toprule
\multirow{2}{*}{\bf{Message Type}} &
\multicolumn{3}{c}{\bf{City to Grid}} & \multicolumn{3}{c}{\bf{Grid to City}} & \multicolumn{3}{c}{\bf{Grid to Grid}}\\
       \cmidrule(lr){2-4}\cmidrule(lr){5-7}\cmidrule(lr){8-10}
 & RMSE & MAE & PCC & RMSE & MAE & PCC & RMSE & MAE & PCC \\ \midrule
 \textsc{Geo} & 126.69 & 38.04 & 0.83 & 107.98 & 35.72& 0.78& 86.62& 29.93& 0.81\\
 \textsc{Flow} & 110.80 & 33.30 & 0.89 & 98.38 & 31.19& 0.83& 87.10& 30.38& 0.81\\
 \textsc{Incl} & 151.41 & 41.12 & 0.74 & 126.88 & 38.36& 0.68& 120.97& 39.56& 0.58\\
 \textsc{Geo $+$ Incl} & 114.40 & 32.93 & 0.89 & 88.51 & 29.48& \underline{0.87}& \underline{84.39}& 29.44& \underline{0.82}\\
 \textsc{Geo $+$ Flow} & 117.38 & \underline{31.67} & 0.88 & 91.23 & \underline{28.83}& \underline{0.87}& 86.46& 29.40& 0.81\\
 \textsc{Flow$+$Geo$+$Incl} & 110.63 & 31.79 & \underline{0.90} & 90.56\ & 29.11& \underline{0.87}& 86.20& 30.32& 0.81\\
 \textsc{HiUrNet$($Flow$+$Incl$)$} & \underline{108.17} & 31.95 & \underline{0.90} & \underline{90.42} & 29.65& \underline{0.87}& 87.84& \underline{29.28}& 0.80\\
\bottomrule
\end{tabular}
\label{table:message_passing_type}
\end{table*}

Although GNN is a popular choice in learning spatial representations, it is not trivial to find the best graph for information exchange. HiUrNet utilized mobility flow and municipal inclusion as the basis of the urban graph. Many previous studies implemented spatial adjacency matrices to formulate graph structures, according to Tobler's first law of geography, which claims that near things are more related than distant ones. However, this rule can be misleading owing to the development of traffic systems that enable long-distance trips. Therefore, we tested different edge combinations in three selected relationships. We generated edges between the central grid and the nearby eight grids for the neighborhood semantics. Table~\ref{table:message_passing_type} indicates the performance comparison of model variants with different combinations of edge types for message passing. HiUrNet achieved almost the best performance over other combinations. It is remarkable to note that because GNNs are likely to suffer from over-smoothing problems~\citep{rusch2023survey}, dense graphs including numerous edges may deteriorate the performance, which can explain that results with all types of edges did not achieve the best performance. Thus, comprehensive constructions of graphs for urban scenarios hold great importance.

\subsubsection{Graph Convolutional Layers}

\begin{table*}
\small
\centering\centering
\setlength{\tabcolsep}{4.2pt}
\caption{Performance for different numbers and types of graph convolutional architectures. The best result is indicated by the underline and the second-to-best result is denoted in bold.} 
\begin{tabular}{p{2cm}ccccccccccc}
\toprule
\multirow{2}{*}{\bf{Conv. Types}} &
\multicolumn{3}{c}{\bf{City to Grid}} & \multicolumn{3}{c}{\bf{Grid to City}} & \multicolumn{3}{c}{\bf{Grid to Grid}} & \multirow{2}{*}{\bf{Training}} & \multirow{2}{*}{\bf{Inferring}}\\
       \cmidrule(lr){2-4}\cmidrule(lr){5-7}\cmidrule(lr){8-10}
 & RMSE & MAE & PCC & RMSE & MAE & PCC & RMSE & MAE & PCC \\ \midrule
 \textsc{1 GAT Layer} & 117.31 & 39.86 & 0.60 & 123.23 & 40.17& 0.70& 139.06& 44.84& 0.81&0.16 s/epoch&0.04 s\\
 \textsc{1 HGT Layer} & 106.55 & 35.05 & 0.69 & 101.08 & 33.79& 0.82& 119.61& 36.65& 0.87&0.35 s/epoch&0.04 s\\
 \textsc{2 GAT Layers} & \textbf{96.16} & 31.87 & \textbf{0.76} & 107.89 & 31.89& 0.82& 150.60& 37.53& 0.78&0.32 s/epoch&0.09 s\\
 \textsc{2 HGT Layers} & 101.69 & 31.11 & 0.72 & 95.07 & \textbf{30.11}& \textbf{0.85}& \underline{110.38}& 32.85& \underline{0.90}&0.33 s/epoch&0.05 s\\
 \textsc{3 GAT Layers} & \multicolumn{11}{c}{CUDA Out of Memory}\\
 \textsc{3 HGT Layers} & \underline{90.19} & \underline{30.19} & \underline{0.80} & \textbf{94.67} & 30.59& \textbf{0.85}& \textbf{111.05}& \textbf{32.59}& \textbf{0.89}&0.91 s/epoch&0.08 s\\
 \textsc{4 GAT Layers} & \multicolumn{11}{c}{CUDA Out of Memory}\\
 \textsc{4 HGT Layers} & 99.39 & \textbf{30.24} & 0.74 & \underline{92.72} & \underline{29.70}& \underline{0.87}& 111.78& \underline{32.37}& \underline{0.90}&1.19 s/epoch&0.10 s\\
\bottomrule
\end{tabular}
\label{table:convolution_type}
\end{table*}

The number of convolutional layers indicates hops of receptive field for GNNs. It determines the target can receive information in one epoch from how many neighboring hops. We also compared model variants with GAT, a commonly used GNN for graph inductive learning. For important hyperparameters, we decided the embedding dimensions as 64 for $\textrm{HiUrNet}_{\textrm{GAT}}$. The number of hidden channels was 128 for $\textrm{HiUrNet}_{\textrm{GAT}}$. We set the number of heads to four for $\textrm{HiUrNet}_{\textrm{GAT}}$. Here, we also provide training and inferring time for the model comparison. Table~\ref{table:convolution_type} shows the comparison of the performance. Note that GAT implementation for homogeneous scenes requires stacking layers according to the number of edge types. Therefore, it consumed lots of memory and caused overflows on the GPU when using more than three GAT layers. However, HGT generally performed better than GAT for the same number of layers and cost less computational resources. The result illustrates that three HGT layers can be the best choice concerning multi-task performance.

\subsection{Sensitivity Analysis for Hyperparameters}

In this subsection, we conduct sensitivity analysis for several critical hyperparameters in this study.

\subsubsection{Analysis for Multi-task Weights}

\begin{figure}
  \centering
  \includegraphics[width=\columnwidth]{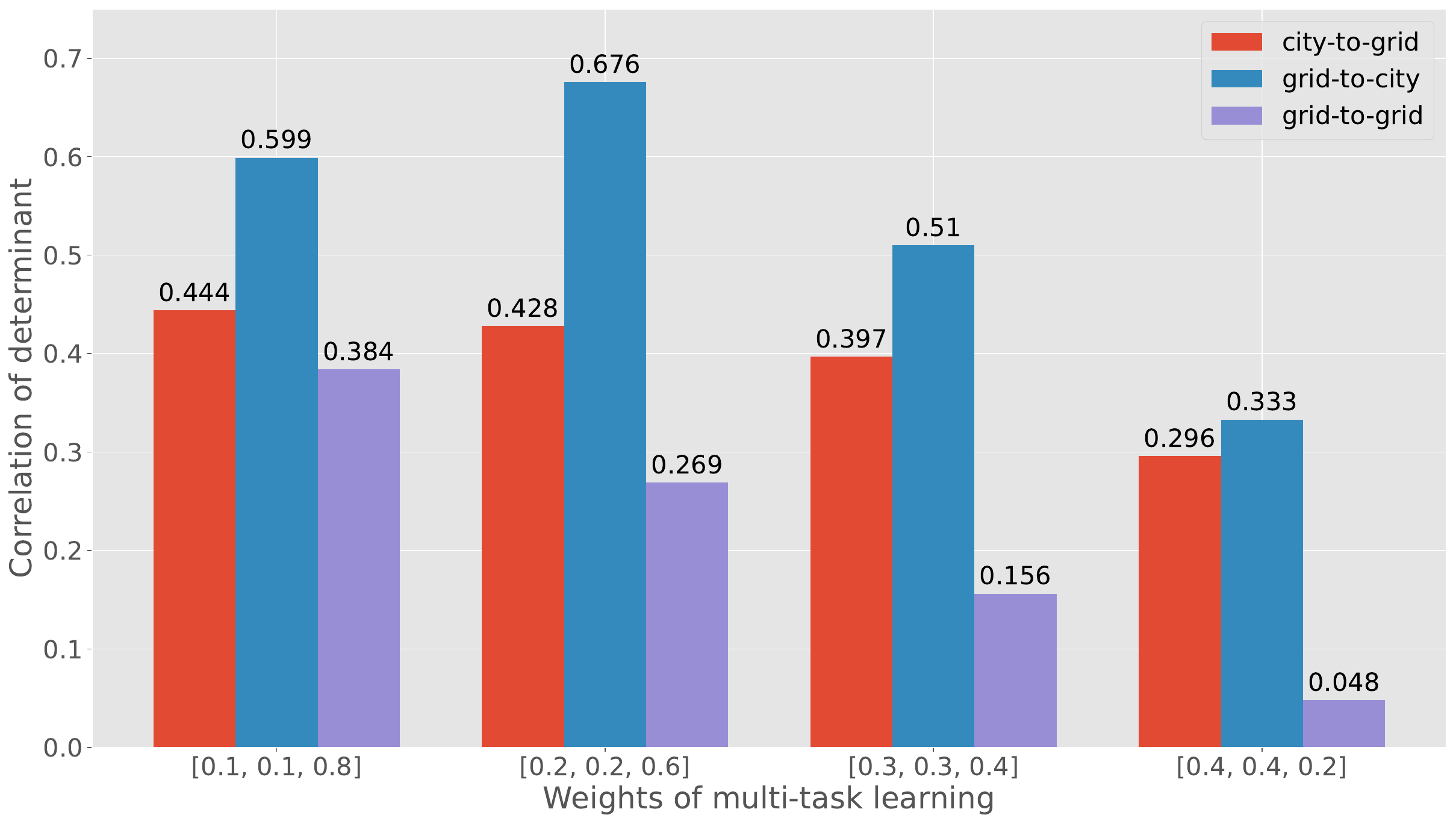}
  \caption{The performance of different combination for multi-task weights.The task weights of $[0.1,0.1,0.8]$ for city-grid, grid-city and grid-grid commuting flows achieve the best performance.} 
  \label{fig:multitask}
\end{figure}

We conducted an analysis to evaluate the influence of distinct combinations of multi-task weights. The prediction tasks for cities–to–grids and grids–to–cities were analogous; thus, we assigned equal weights to these two subtasks. We tested several groups of weight combinations and reported the correlation of determinants as metrics of the performance. The results are indicated in figure~\ref{fig:multitask}. The task weights of $[0.1,0.1,0.8]$ achieve the best performance, which illustrates that the task of grid-to-grid commuting flow plays a more significant role than the other sub tasks. The number of these grid-to-grid commuting pairs is massive. Moreover, grids can have direct effect on cities through inclusion relationship and thus determined the accuracy of the other types of commuting trips simultaneously.

\subsubsection{Embedding Size}

\begin{figure}[htb]
    \centering
\begin{subfigure}{0.49\columnwidth}
  \includegraphics[width=\linewidth]{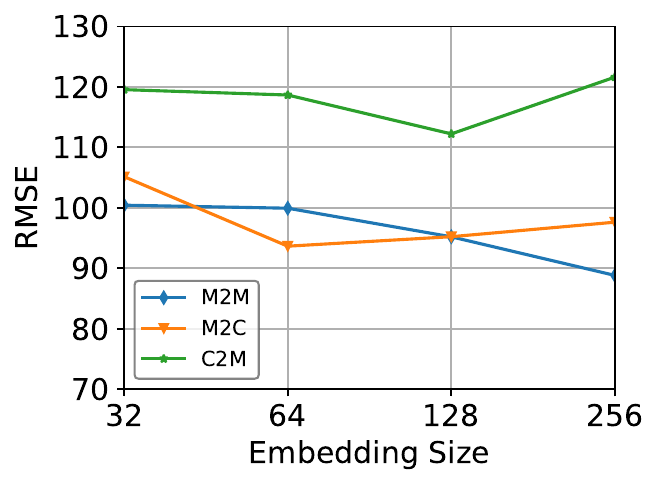}
  \label{fig:embedding_size_rmse}
\end{subfigure}\hfil 
\begin{subfigure}{0.49\columnwidth}
  \includegraphics[width=\linewidth]{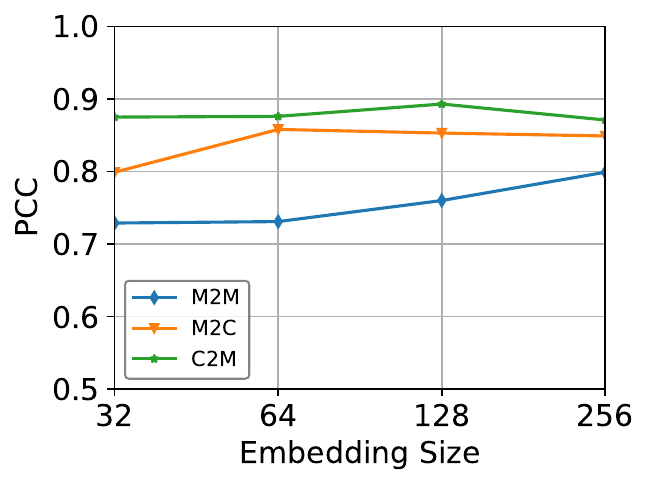}
  \label{fig:embedding_size_pcc}
\end{subfigure}
\caption{The sensitivity analysis for the embedding size.}
\label{fig:embedding_size}
\end{figure}

The embedding size can be a critical factor for representation models to aggregate information from input data. The candidate embedding size is selected from $[32,64,128,256]$. Figure~\ref{fig:embedding_size} denotes the effect of different sizes for urban unit representation. We utilized RMSE and PCC as metrics to determine the appropriate choice. Considering the best and second-to-best performance in three tasks, we choose 128 as the final embedding size.

\section{Conclusion}

This study proposes a HiUrNet model to describe urban areas on a large scale. We conducted multi-task learning to balance the tradeoff between the prediction of different types of commuting flows. This aligns with real-world requirements, striking a balance between the need for large-scale predictions and high-resolution insights into human mobility. We also determined urban knowledge using the topological structure from the mobility and municipal aspects. The results indicate that the model can capture the patterns of massive multi-resolution urban flows.

The proposed model can derive hierarchical embeddings throughout the open data for ease of use. Significantly, the government can apply the proposed model on a large urban scale to analyze macroscopic measures based on the human mobility. Because the model learns a mapping function from grid indicators to grid embeddings and a summary function from grid embeddings to city embeddings, it can avoid the collection of data at a high-cost, which is crucial for saving time and budget.

In future work, we will consider more types of spatial units and the relationships between units at different levels to comprehensively illustrate urban semantics.

\begin{acks}
This work was partially supported by JSPS KAKENHI Grant Numbers JP21K14260. We appreciate SoftBank Group Corporation for providing the mobility data used in this research. 
\end{acks}

\bibliographystyle{ACM-Reference-Format}
\bibliography{reference}

\appendix

\end{document}